%% file: paper.tex
\documentclass{article}

\usepackage{microtype}
\usepackage{graphicx}
\usepackage{booktabs} % for professional tables

\usepackage{amsmath}
\usepackage{amssymb}
\usepackage{amsfonts}
\usepackage{caption}
\usepackage{subcaption}
\usepackage{color}
\newcommand{\bx}{{X}}
\newcommand{\app}{{Appendix}}

\usepackage{hyperref}

\newtheorem{theorem}{Theorem}
\newtheorem{proposition}[theorem]{Proposition}
\newtheorem{corollary}{Corollary}
\newcommand{\BlackBox}{\rule{1.5ex}{1.5ex}}  % end of proof
\newenvironment{proof}{\par\noindent{\bf Proof\ }}{\hfill\BlackBox\\[2mm]}
\usepackage[accepted]{icml2019}

\input{math_commands.tex}

\icmltitlerunning{Curriculum Learning in Deep Networks}

\begin{document}

\twocolumn[
\icmltitle{On The Power of Curriculum Learning in Training Deep Networks}



\icmlsetsymbol{equal}{*}

\begin{icmlauthorlist}
\icmlauthor{Guy Hacohen}{huji,elsc}
\icmlauthor{Daphna Weinshall}{huji}
\end{icmlauthorlist}

\icmlaffiliation{huji}{School of Computer Science and Engineering, The Hebrew University of Jerusalem, Jerusalem 91904, Israel}
\icmlaffiliation{elsc}{Edmond and Lily Safra Center for Brain Sciences, The Hebrew University of Jerusalem, Jerusalem 91904, Israel}

\icmlcorrespondingauthor{Guy Hacohen}{guy.hacohen@mail.huji.ac.il}

% the ``keywords" metadata in the PDF but will not be shown in the document
\icmlkeywords{Curriculum learning, transfer learning, deep networks}
\vskip 0.3in
]



\printAffiliationsAndNotice{}  % leave blank if no need to mention equal contribution

\begin{abstract} 
% Training neural networks is traditionally done by sequentially providing random mini-batches sampled uniformly from the entire dataset. In our work, we show that sampling mini-batches non-uniformly can both enhance the speed of learning and improve the final accuracy of the trained network. Specifically, we decompose the problem using the principles of curriculum learning: first, we sort the data by some difficulty measure; second, we sample mini-batches with a gradually increasing level of difficulty. We focus on CNNs trained on image recognition. Initially, we define the difficulty of a training image using transfer learning from some competitive ``teacher" network trained on the Imagenet dataset, showing improvement in learning speed and final performance for both small and competitive networks, using the CIFAR-10 and the CIFAR-100 datasets. We then suggest a bootstrap alternative to evaluate the difficulty of points using the same network without relying on a ``teacher" network, thus increasing the applicability of our suggested method. We compare this approach to a related version of Self-Paced Learning, showing that our method benefits learning while SPL impairs~it.\guy{need to revise to match the intro. last thing to do}

Training neural networks is traditionally done by providing a sequence of random mini-batches sampled uniformly from the entire training data. In this work, we analyze the effect of curriculum learning, which involves the non-uniform sampling of mini-batches, on the training of deep networks, and specifically CNNs trained for image recognition. To employ curriculum learning, the training algorithm must resolve 2 problems: (i) sort the training examples by difficulty; (ii) compute a series of mini-batches that exhibit an increasing level of difficulty. We address challenge (i) using two methods: transfer learning from some competitive ``teacher" network, and bootstrapping. In our empirical evaluation, both methods show similar benefits in terms of increased learning speed and improved final performance on test data. We address challenge (ii) by investigating different pacing functions to guide the sampling. The empirical investigation includes a variety of network architectures, using images from CIFAR-10, CIFAR-100 and subsets of ImageNet. We conclude with a novel theoretical analysis of curriculum learning, where we show how it effectively modifies the optimization landscape. We then define the concept of an ideal curriculum, and show that under mild conditions it does not change the corresponding global minimum of the optimization function.

%non-uniformly can both enhance the speed of learning and improve the final accuracy of the trained network. Specifically, we decompose the problem using the principles of curriculum learning: first, we sort the data by some difficulty measure; second, we sample mini-batches with a gradually increasing level of difficulty. We focus on CNNs trained on image recognition. Initially, we define the difficulty of a training image using transfer learning from some competitive ``teacher" network trained on the Imagenet dataset, showing improvement in learning speed and final performance for both small and competitive networks, using CIFAR-10, CIFAR-100 and subset of ImageNet. We then suggest a bootstrap alternative to evaluate the difficulty of points using the trained network without external ``teacher" network or additional supervised data, increasing the applicability of our suggested method. We conclude by empirical and theoretical analysis to explain the effects of curriculum learning on deep neural networks. \guy{i re-wrote the abstract. as for now, it emphasize the first part of our work, while after the reviews, we might want to shift that emphasis to the second part. should i re-write it to emphasize something else? or should i leave it as it is?}

\end{abstract}

\section{Introduction}
\label{sec:intro}

%When teaching complex tasks to humans and animals, often they cannot be grasped by the learners (or ``students") immediately, and need to be broken down into simpler problems. Therefore, 
In order to teach complex tasks, teachers are often required to create a curriculum. A curriculum imposes some order on the concepts that constitute the final task, an order which typically reflects their complexity. The student is then gradually introduced to these concepts by increasing complexity, in order to allow her to exploit previously learned concepts and thus ease the abstraction of new ones. But the use of a curriculum is not limited to complex tasks. When teaching a binary classification task, for example, teachers tend to present typical examples first, followed by the more ambiguous examples \cite{avrahami1997teaching}. 

In many traditional machine learning paradigms, a target function is estimated by a learner (the ``student") using a set of training labeled examples (provided by the ``teacher"). The field of curriculum learning (CL), which is motivated by the idea of a curriculum in human learning, attempts at imposing some structure on the training set. Such structure essentially relies on a notion of ``easy" and ``hard" examples, and utilizes this distinction in order to teach the learner how to generalize easier examples before harder ones. Empirically, the use of CL has been shown to accelerate and improve the learning process \citep[e.g.][]{selfridge1985training,bengio2009curriculum} in many machine learning paradigms.  
%\guy{i guess i can add more refs here. but i think this will do for now}. 

When establishing a curriculum for human students, teachers need to address two challenges: (i) Arrange the material in a way that reflects difficulty or complexity, %so that the abstraction of simple ideas can help the student grasp more complex ones \citep{hunkins2016curriculum}, 
a knowledge which goes beyond what is available in the training set in most machine learning paradigms. (ii) Attend to the pace by which the material is presented -- going over the simple ideas too fast may lead to more confusion than benefit, while moving along too slowly may lead to boredom and unproductive learning \citep{hunkins2016curriculum}. In this paper, we study how these principles can be beneficial when the learner is a neural network. 

In order to address the first challenge, \citet{weinshallcurriculum} introduced the idea of curriculum learning by transfer. The idea is to sort the training examples based on the performance of a pre-trained network on a larger dataset, fine-tuned to the dataset at hand. This approach was shown to improve both the speed of convergence and final accuracy for convolutional neural networks, while not requiring the manual labeling of training data by difficulty. 

In our work, we address both challenges. We begin by %formalizing and generalizing what was implicitly done in \citet{weinshallcurriculum}, 
decomposing CL into two separate - but closely related - sub-tasks and their corresponding functions. The first, termed \emph{scoring function}, determines the ``difficulty" or ``complexity" of each example in the data. The \emph{scoring function} makes it possible to sort the training examples by difficulty, and present to the network the easier (and presumably simpler) examples first. %The underlying assumption is that generalization from the easier examples can simplify the learning of harder examples in the data. 
Scoring is done based on transfer learning as in \citet{weinshallcurriculum}, or bootstrapping as explained below.
The second function, termed \emph{pacing function}, determines the pace by which data is presented to the network. The pace may depend on both the data itself and the learner.
%: as in the case of human students, some network architectures will need a slower pace for a large variety of problems, while others can learn to generalize from the same data using faster pace.

We show that the use of different pacing functions can indirectly affect hyper-parameters of the neural network, reminiscent of increased learning rate. As \citet{weinshallcurriculum} did not employ parameter tuning in their empirical study, the improvement they report might be explained by the use of inappropriate learning rates. As part of our work, we repeat their experimental paradigm while optimizing the network's hyper-parameters (with and without cross-validation), showing improvement in both the speed of convergence and final accuracy in a more reliable way. We then extend these results and report experiments on larger datasets and architectures, which are more commonly used as benchmarks.
% We analyze different of pacing functions, and suggest a new scoring function which reach similar qualitative results without the need of additional human supervision. Finally, we perform a theoretical analysis might explain some of the empirical improvements we see.

We continue by analyzing several \emph{scoring} and \emph{pacing} functions, investigating their inter-dependency and presenting ways to combine them in order to achieve faster learning and better generalization. The main challenge is, arguably, how to obtain an effective \emph{scoring function} without additional labeling of the data. To this end we investigate two approaches, each providing a different estimator for the target \emph{scoring function}: (i) Knowledge transfer as in  \citep{weinshallcurriculum}, based on transfer learning from networks trained on the large and versatile Imagenet dataset. (ii) Bootstrapping based on self-tutoring - we train the network without curriculum, then use the resulting classifier to rank the training data in order to train the same network again from scratch. Unlike curriculum by transfer, bootstrapping does not require access to any additional resources. Both \emph{scoring functions} maintain the required property that prefers points with a lower loss with respect to the target hypothesis. The aforementioned functions are shown in Section~\ref{sec:empirical} to speed up learning and improve the generalization of CNNs.

We investigate three \emph{pacing functions}. (i) \emph{Fixed exponential pacing} presents the learner initially with a small percentage of the data, increasing the amount exponentially every fixed number of learning iterations. (ii) \emph{Varied exponential pacing} allows the number of iterations in each step to vary as well. (iii) \emph{Single-step pacing} is a simplified version of the first protocol, where mini-batches are initially sampled from the easiest examples (a fixed fraction), and then from the whole data as usual. In our empirical setup, the three functions have comparable performance.

In Section~\ref{sec:theory} we conclude with a theoretical analysis of the effects of curriculum learning on the objective function of neural networks. We show that curriculum learning modifies the optimization landscape, making it steeper while maintaining the same global minimum of the original problem. This analysis provides a framework by which apparently conflicting heuristics for the dynamic sampling of training points can coexist and be beneficial, including SPL, boosting and hard data mining, as discussed under \emph{previous work}.
%Assuming the hypothesis around the minimum are correlative with it, such decrease can partially explain the faster convergence in curriculum learning.

\vspace{-0.5cm}
\paragraph{Previous work.}
Imposing a curriculum in order to speed up learning is widely used in the context of human learning and animal training \citep{skinner1958reinforcement, pavlov2010conditioned, krueger2009flexible}. In many application areas, it is a common practice to introduce concepts in ascending order of difficulty, as judged by either the human teacher or in a problem dependent manner \citep[e.g.][]{murphy2008search,zaremba2014learning,amodei2016deep}. With the rebirth of deep learning and its emerging role as a powerful learning paradigm in many applications, the use of CL to control the order by which examples are presented to neural networks during training is receiving increased attention \citep{graves2016hybrid, graves2017automated, florensa2017reverse}. 

In a closely related line of work, a pair of teacher and student networks are trained simultaneously, where mini-batches for the student network are sampled dynamically by the teacher, based on the student's output at each time point. As opposed to our method, here the curriculum is based on the current hypothesis of the student, while achieving improved performance for corrupted \citep{jiang2018mentornet} or smaller \citep{fan2018learning} datasets. Improvement in generalization over the original dataset has not been shown.

In some machine learning paradigms, which are related to CL but differ from it in an essential manner, mini-batches are likewise sampled dynamically. Specifically, in Self-Paced Learning \citep[SPL-][]{kumar2010self}, boosting \citep{freund1996experiments}, hard example mining \citep{shrivastava2016training} and even active learning \citep{schein2007active}, mini-batches are sampled at each time point based on the ranking of the training examples by their difficulty with respect to the \textbf{current hypothesis} of the model. Thus they differ from CL, which relies on the ranking of training points by their difficulty with respect to some \textbf{target hypothesis}. 

Confusingly, based on the same ephemeral ranking, SPL advocates the use of easier training examples first, while the other approaches prefer to use the harder examples. Still, all approaches show benefit under different empirical settings \citep{chang2017active, zhang2017spftn}. This discrepancy is analyzed in \citet{weinshallcurriculum}, where it is shown that while it is beneficial to prefer easier points with respect to the target hypothesis as advocated by CL, it is at the same time beneficial to prefer the more difficult points with respect to the current hypothesis in agreement with hard data mining and boosting, but contrary to SPL. In contrast, our theoretical analysis (Section~\ref{sec:theory}) is consistent with both heuristics being beneficial under different circumstances.

% In related areas such as robotics, it is common practice to introduce easier concepts to a learner followed by more complex concepts in ascending order of difficulty, where difficulty is judged by the human operator \citep{murphy2008search}. In the evaluation of computer programs \citep{zaremba2014learning} or machine translation \citep{amodei2016deep}, problem-dependent automatic criteria for the evaluation of task difficulty have been used.

% \textbf{Our contribution}, with respect to this previous work, is to provide a formal definition of CL algorithms by way of 2 functions for scoring and pacing, analyze and comparatively evaluate these functions, and show how CL can benefit learning in CNNs even without human supervision about the ranking of examples by difficulty and in a problem-free manner.

%\guy{target: page and a half until this point}

\section{Curriculum Learning}

Curriculum learning as investigated here deals with the question of \emph{how to use prior knowledge about the difficulty of the training examples}, in order to sample each mini-batch non-uniformly and thus boost the rate of learning and the accuracy of the final classifier. The paradigm of CL is based on the intuition that it helps the learning process when the learner is presented with simple concepts first. 

\subsection{Notations and Definitions}

Let $\sX = \left\{X_i\right \}_{i=1}^N = \left\{(\vx_i,y_i)\right \}_{i=1}^N$ denote the data, where $\vx_i\in\sR^d$ denotes a single data point and $y_i\in [K]$ its corresponding label. Let $h_\vartheta:\sR^d\rightarrow [K]$ denote the target classifier (or learner), and mini-batch $\sB\subseteq \sX$ denote a subset of $\sX$. In the most common training procedure, which is a robust variant of Stochastic Gradient Descent (SGD), $h_\vartheta$ is trained sequentially when given as input a sequence of mini-batches $\left[\sB_1,...,\sB_M\right]$ \citep{shalev2014understanding}. The common approach -- denoted henceforth \emph{vanilla} -- samples each mini-batch $\sB_i$ uniformly from $\sX$.
Both in the common approach and in our work, the size of each mini-batch remains constant, to be considered as a hyper-parameter defining the learner.

We measure the difficulty of example $(\vx_i, y_i)$ by its minimal loss with respect to the set of optimal hypotheses under consideration. We define a \emph{scoring function} to be any function $f:\sX\rightarrow\sR$, and say that example $(\vx_i, y_i)$ is more difficult than example $(\vx_j, y_j)$ if $f\left(\vx_i, y_i\right)>f\left(\vx_j, y_j\right)$. Choosing $f$ is the main challenge of CL, as it encodes the prior knowledge of the teacher.

We define a \emph{pacing function} $g_{\vartheta}:[M]\rightarrow [N]$, which may depend on the learner $h_\vartheta$. The \emph{pacing function} is used to determine a sequence of subsets $\sX_{1}^{'},...,\sX_{M}^{'}\subseteq\sX$, of size $|\sX_{i}^{'}|=g_{\vartheta}(i)$, from which $\{\sB_i\}_{i=1}^M$ are sampled uniformly. In CL the $i$-th subset $\sX_{i}^{'}$ includes the first $g_{\vartheta}(i)$ elements of the training data when sorted by the \emph{scoring function} $f$ in ascending order. Although the choice of the subset can be encoded in the distribution each $\sB_i$ is sampled from, adding a \emph{pacing function} simplifies the exposition and analysis.

\subsection{Curriculum Learning Method}

Together, each \emph{scoring function $f$} and \emph{pacing function} $g_{\vartheta}$ define a curriculum. Any learning algorithm which uses the ensuing sequence $[\sB_i]_{i=1}^M$ is a \textbf{curriculum learning algorithm}. We note that in order to avoid bias when picking a subset of the $N$ examples for some $N$, it is important to keep the sample balanced with the same number of examples from each class as in the training set. Pseudo-code for the CL algorithm is given in Alg.~\ref{alg: curriculum}.

\begin{algorithm}[tb]
   \caption{Curriculum learning method}
   \label{alg: curriculum}
\begin{algorithmic}
   \STATE {\bfseries Input:} \emph{pacing function} $g_{\vartheta}$, \emph{scoring function} $f$, data $\sX$.
   \STATE {\bfseries Output:} sequence of mini-batches $\left[\sB_{1}^{'},...,\sB_{M}^{'}\right]$.
   \STATE sort $\sX$ according to $f$, in ascending order\;
   \STATE $result \leftarrow[]$\;
   \FORALL{$i=1,...,M$}
     \STATE $size \leftarrow g_{\vartheta}(i)$\;
     \STATE $\sX_{i}^{'}\leftarrow\sX\left[1,...,size\right]$\;
     \STATE uniformly sample $\sB_{i}^{'}$ from $\sX^{'}$\;
     \STATE append $\sB_{i}^{'}$ to $result$\;
   \ENDFOR
   \STATE {\bfseries return} $result$\;
\end{algorithmic}
\end{algorithm}

In order to narrow down the specific effects of using a \emph{scoring function} based on ascending difficulty level, we examine two control conditions. Specifically, we define 2 additional \emph{scoring functions} and corresponding algorithms: (i) The \textbf{anti-curriculum algorithm} uses the \emph{scoring function} $f'=-f$, where the training examples are sorted in  \emph{descending order} of difficulty; thus harder examples are sampled before easier ones. (ii) The \textbf{random-curriculum algorithm} (henceforth denoted \textbf{random}) uses a \emph{scoring function} where the training examples are randomly scored.

%\subsection{Curriculum by transfer and self-taught: definitions}

%The main hurdle to actually using CL is the scoring function - how can we score the difficulty of the training examples without access to the optimal hypothesis, which defines a point's difficulty level? Here we investigate two directions in order to approximate this scoring function. In \emph{curriculum by transfer}, the \emph{scoring function} is obtained from a bigger network trained on a larger dataset. % - the Inception network. 
%In \emph{curriculum by self tutoring}, the \emph{scoring function} is obtained from the same network, pre-trained on the same data without CL, followed by CL from scratch. \guy{mihgt be fine, but we did say these ideas in the introduction. should we remove 1?}

\subsection{Scoring and Pacing Functions}
\label{sec:scoring-pacing-def}

We evaluate two \emph{scoring functions}: (i) \emph{Transfer scoring function}, computed as follows: First, we take the Inception network \citep{szegedy2016rethinking} pre-trained on the ImageNet dataset \citep{deng2009imagenet} and run each training image through it, using the activation levels of its penultimate layer as a feature vector \citep{caruana1995learning}. Second, we use these features to train a classifier and use its confidence score as the \emph{scoring function} for each image\footnote{Similar results are obtained when using different confidence scores (e.g, the classifier's margin), different classifiers (e.g, linear SVM), and different teacher networks (e.g, VGG-16 \citep{simonyan2014very} or ResNet \citep{he2016deep}), see \app~\ref{appendix:transfer_networks_comparison}.}. (ii) \emph{Self-taught scoring function}, computed as follows: First, we train the network using uniformly sampled mini-batches (the \emph{vanilla} method). Second, we compute this network's confidence score for each image to define a \emph{scoring function}\footnote{\emph{Self-taught} can be used repeatedly, see \app~\ref{app:self-bootsrapping}.}.

\begin{figure}[ht!]
    \centering
    \begin{subfigure}[t]{0.5\textwidth}
        \includegraphics[width=\textwidth]{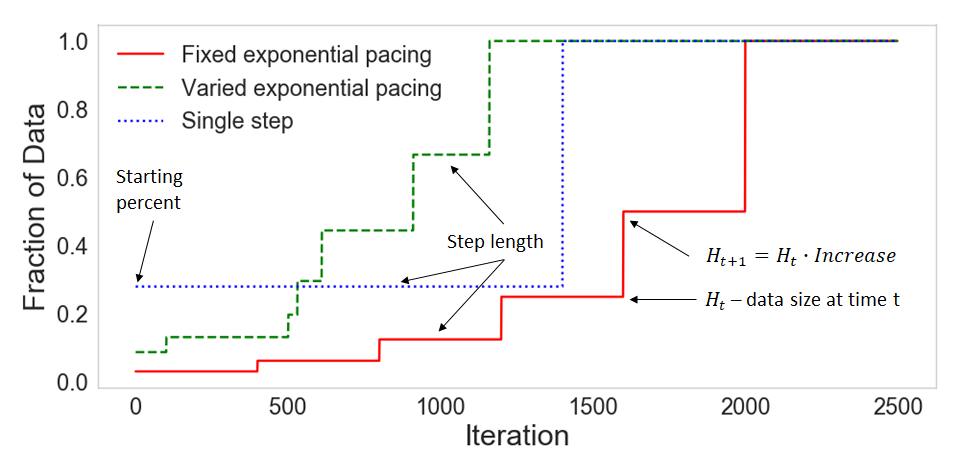}
        %\subcaption{Our results}
        %\label{fig:problem1-ours}
    \end{subfigure}
    \vspace*{-8mm}
    \caption{Illustration of the 3 \emph{pacing functions} used below, showing the different hyper-parameters that define each of them (see text). %In red, we see the \emph{fixed exponential pacing function}, which has fixed \textsl{step\_length}, and exponentially increasing \textsl{step\_height}. In green, we see the \emph{varied exponential pacing function}, which has varied \textsl{step\_length} and exponentially increasing step size. In blue, we see the \emph{single step pacing function}, which is a step function. 
The values of the hyper-parameters used in this illustration were chosen arbitrarily, for illustration only.}
    \label{fig:pacing_functions}
\end{figure}

The \emph{pacing function} can be any function $g_{\vartheta}:[M]\rightarrow [N]$. However, we limit ourselves to monotonically increasing functions so that the likelihood of the easier examples can only decrease. For simplicity, $g_{\vartheta}$ is further limited to staircase functions. Thus each \emph{pacing function} is defined by the following hyper-parameters, where \textsl{step} denotes all the learning iterations during which $g_{\vartheta}$ remains constant: \textsl{step\_length} - the number of iterations in each \textsl{step}; \textsl{increase} - an exponential factor used to increase the size of the data used for sampling mini-batches in each \textsl{step}; \textsl{starting\_percent} - the fraction of the data in the initial \textsl{step}. An illustration of these parameters can be seen in Fig.~\ref{fig:pacing_functions}.

We evaluate three \emph{pacing functions}: (i) \emph{Fixed exponential pacing} has a fixed \textsl{step\_length}, and exponentially increasing data size in each \textsl{step}. Formally,  it is given by:
\begin{equation*}
g_{\vartheta}(i) = \min\left(starting\_percent\cdot inc^{\floor{\frac{i}{step\_length}}},1\right)\cdot N
\end{equation*} (ii) \emph{Varied exponential pacing} is the same as (i), while allowing \textsl{step\_length} to vary. This method adds additional hyper-parameters, but removes the need to re-tune the learning rate parameters (see \app~\ref{app: Varied_exponnetial}). (iii) \emph{Single step pacing} entails the simplification of the staircase function into a step function, resulting in fewer hyper-parameters. Formally:%, it is given by:
\begin{equation*}
g_{\vartheta}(i) = starting\_percent^{\1{\left[i<step\_length\right]}}\cdot N
\end{equation*}

\section{Empirical Evaluation}
\label{sec:empirical}

\paragraph{Methodology.\footnote{All the code used in the paper is available at  https://github.com/GuyHacohen/curriculum\_learning}}
We define 6 test cases: \textbf{Case 1} replicates the experimental design described in \citep{weinshallcurriculum}, by using the same dataset and network architecture. The dataset is the ``small mammals" super-class\footnote{Other super-classes achieve similar results, see \app~\ref{app:curriculum-subset-0}.} of CIFAR-100 \citep{krizhevsky2009learning} - a subset of 3000 images from CIFAR-100, divided into 5 classes.
% containing a subset of 3000 images from CIFAR-100, divided into 5 classes of small mammals (hamster, mouse, rabbit, shrew, squirrel). Each class contains 500 training images and 100 test images.
The neural network is a moderate size hand-crafted convolutional network, whose architecture details can be found in \app~\ref{app:architecture}.
%The results of \citep{weinshallcurriculum} for case 1 are depicted in Fig.~\ref{fig:gad}.
%\begin{figure}[ht!]
%\begin{center}
%\framebox[4.0in]{$\;$}
%\includegraphics[width=0.6\textwidth]{gad_learning_acceleration_mod.png}
%\end{center}
%\caption{Curriculum results from \cite{weinshallcurriculum}. Learning curves for case 1. The curriculum test case (in green) reaches higher accuracy in the early epochs compared to the vanilla test case (in blue). Both the random and the anti-curriculum test cases (yellow and red lines respectively) reach lower final accuracy than the vanilla test case.}
%\label{fig:gad}
%\end{figure}
\textbf{Cases 2} and \textbf{3} adopt the same architecture while extending the datasets to the entire CIFAR-10 and CIFAR-100 datasets. % \citep{krizhevsky2009learning}.
\textbf{Cases 4} and \textbf{5} use a well known public-domain VGG-based architecture\footnote{Code available at https://github.com/geifmany/cifar-vgg.} \citep{simonyan2014very, liu2015very}, to classify the CIFAR-10 and CIFAR-100 datasets. \textbf{Case 6} adopts the same architecture as cases 1-3, trained with a subset of 7 classes of cats (see \app~\ref{app: imagenet-cats}) from the ImageNet dataset.

%\subsection{Curriculum learning by transfer}

%Analyzing the results of \cite{weinshallcurriculum} (Fig.~\ref{fig:problem1-gad}), one can see that the learning curve of the random test case is generally lower than the vanilla's learning curve. This difference suggests that their choice of pacing function impedes the network's learning ability, as the pacing function is the only difference between these cases. As the curriculum test case only differs from the random test case in the choice of the \emph{scoring function} $f$, any additional acceleration achieved by $f$ will need to first overcome the negative impact of such pacing function. Hence, we hypothesize that improving the random test case, by overcoming the negative effects of the pacing function, might also improve the corresponding curriculum test case.

\paragraph{Hyper-parameter tuning.}

As in all empirical studies involving deep learning, the results are quite sensitive to the values of the hyper-parameters, hence parameter tuning is required. %Issues related to how we achieved a fair comparison between the different conditions are discussed in \app~\ref{app:fair}. 
In practice, in order to reduce the computation time of parameter tuning, in the curriculum conditions, we performed first grid-search on the curriculum hyper-parameters, followed by a second grid-search on the learning rate parameters, thus avoiding the need to tune a large number of parameters at once. In the non-curriculum conditions, a full 1-stage grid-search was performed. In addition, we varied only the first 2 \textsl{step\_length} instances in the \emph{varied exponential pacing} condition. Accordingly, \emph{fixed exponential pacing}, \emph{varied exponential pacing} and \emph{single step pacing} define 3, 5 and 2 new hyper-parameters respectively, henceforth called the \emph{pacing} hyper-parameters. 
% \emph{Fixed exponential pacing}, \emph{varied exponential pacing} and \emph{single step pacing} define 3, $\#steps + 2$ and 2 new hyper-parameters respectively. In practice, in the case of \emph{varied exponential pacing}, we varied only the first 2 steps, resulting in $\#steps=3$ to a total of 5 new hyper-parameters.\daphna{can't understand this last sentence} These parameters are referred to as the pacing hyper-parameters.\daphna{the last 2 sentences are way overkill, no need to get to this level of details in the methodology; can you shorten it to one sentence?}

With CL, the use of a \emph{pacing function} affects the optimal values of other hyper-parameters, the learning rate in particular. Specifically, the \emph{pacing function} significantly reduces the size of the dataset at the beginning of the learning, %%More specifically, in SGD typically the learning rate is gradually decreased, and its optimal value in each epoch depends in part upon the size of the mini-batch and the size of the training set. When the training set size is effectively reduced by the \emph{pacing function}, this
which has the concomitant effect of increasing the effective learning rate at that time. As a result, for a fair comparison, when using the \emph{fixed exponential} or the \emph{single step pacing functions}, the learning rate must be tuned separately for every test condition. This tuning is missing in \citet{weinshallcurriculum}, whose results may therefore be tainted by their arbitrary choice of learning rate. Using the \emph{varied exponential pacing function} can overcome the need for learning rate re-tuning, while adding 2 hyper-parameters (see \app~\ref{app: Varied_exponnetial}).

As traditionally done \citep[e.g][]{simonyan2014very,szegedy2016rethinking,he2016deep}, we set an initial learning rate and decrease it exponentially every fixed number of iterations. This method gives rise to 3 learning rate hyper-parameters which require tuning: (i) the initial learning rate; (ii) the factor by which the learning rate is decreased; (iii) the length of each step with constant learning rate\footnote{Other tuning methods achieve similar results, see \app~\ref{app:learning-rate-tuning}.}.

\paragraph{Cross-validation.}
In grid search, parameters are chosen based on performance on the test set. To avoid contamination of the conclusions, all results were cross-validated, wherein the hyper-parameters are chosen based on performance on a validation set before being used on the test set. For more details on the steps we took to ensure a fair comparison when employing a grid search, see \app~\ref{app:fair}.

% The need of tuning the learning rate separately for the \emph{vanilla} and the \emph{fixed exponential} pacing can be avoided by allowing the \textsl{step\_length} to change in each step, resulting in the \emph{varied exponential pacing}. For more details, see \app. \guy{add to \app}
% When \emph{varied exponential pacing} is used, varying \textsl{step\_length} has the opposite concomitant effect on the learning rate, as it determines the number of mini-batch samples in each \textsl{step}. Effective tuning of this parameter can make the additional tuning of parameters affecting the learning rate redundant. In practice, in order to reach the improvement achieved by the \emph{fixed exponential pacing}, we decrease the corresponding  learning rate parameters used in the \emph{vanilla} condition by some small factor\footnote{In the results reported below we used a reduction of $10\%$, with similar behavior for other nearby choices.}.

\subsection{Curriculum by Transfer}

\begin{figure}[ht!]
    \centering
    \begin{subfigure}[t]{0.5\textwidth}
        \includegraphics[width=\textwidth]{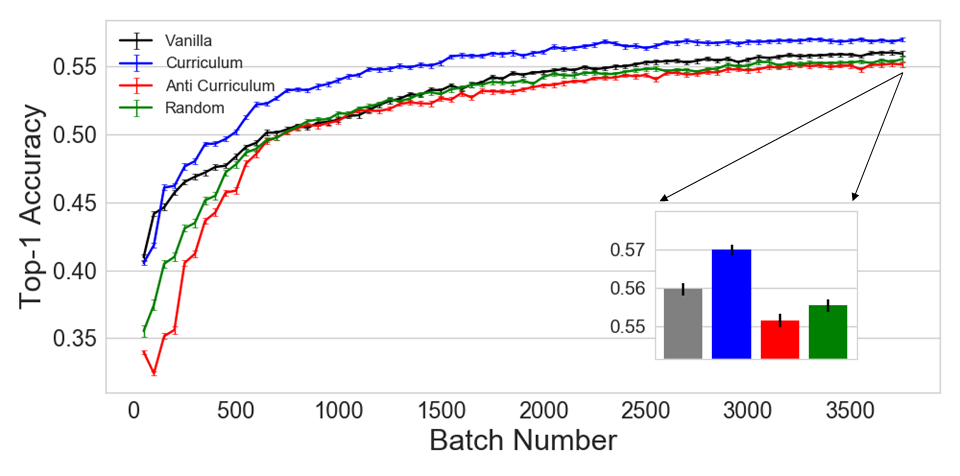}
        %\caption{Our results}
        %\label{fig:problem1-ours}
    \end{subfigure}
    \vspace*{-8mm}
    \caption{Results in \textbf{case 1}, with Inception-based \emph{transfer scoring function} and \emph{fixed exponential pacing function}. Inset: bars indicating the average final accuracy in each condition over the last few iterations. Error bars indicate the STE (STandard Error of the mean) after 50 repetitions. The curriculum method (in blue) reaches higher accuracy faster, and converges to a better solution.}\label{fig:[problem1]}
    
\end{figure}

\paragraph{Case 1:}
A moderate size network is trained to distinguish 5 classes from CIFAR-100, which are members of the same super-class as defined in the original dataset. Results are shown in Fig.~\ref{fig:[problem1]}. Curriculum learning is  clearly and significantly beneficial - learning starts faster, and converges to a better solution. We observe that the performance of CL with a random \emph{scoring function} is similar to \emph{vanilla}, indicating that the main reason for the improvement achieved by CL is due to its beneficial \emph{transfer scoring function}. In fact, although tuned separately, the learning rate hyper-parameters for both the \emph{random} and the \emph{curriculum} test conditions are very similar, confirming that the improved performance is due to the use of an effective \emph{transfer scoring function}. 

To check the robustness of these results, we repeated the same empirical evaluation using different super-classes of CIFAR-100, with similar results (see \app~\ref{app:curriculum-subset-0}). 
%Aquatic mammals and people super-classes were picked due to their low vanilla test accuracy. Insects super-class was picked at random.
Interestingly, we note that the observed advantage of CL is more significant when the task is more difficult (i.e. lower \emph{vanilla} test accuracy). The reason may be that in easier problems there is a sufficient number of easy examples in each mini-batch even without CL. Although the results reported here are based on transfer from the Inception network, we are able to obtain the same results using scoring functions based on transfer learning from other large networks, including VGG-16 and ResNet, as shown in \app~\ref{appendix:transfer_networks_comparison}.

\begin{figure}[ht!]
    \centering
%     \begin{subfigure}[b]{0.495\textwidth}
%         \includegraphics[width=\textwidth]{graphs/problem2_curriculum_vanilla_cifar10_36epochs.png}
%         \caption{\textbf{Case 2}: CIFAR-10}
%         \label{fig:problem2-cifar10-curves}
%     \end{subfigure}
    \begin{subfigure}[b]{0.495\textwidth}
        \includegraphics[width=\textwidth]{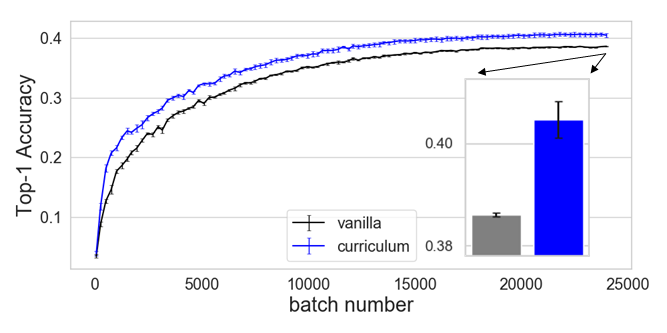}
%         \caption{\textbf{Case 3}: CIFAR-100}
    \end{subfigure}
    \label{fig:problem3-cifar100-curves}
    \vspace*{-8mm}
    \caption{Results in \textbf{case 3}, CIFAR-100 dataset, with Inception-based \emph{transfer scoring function} and \emph{fixed exponential pacing function}. Inset: zoom-in on the final iterations, for better visualization. Error bars show STE after 5 repetitions.}\label{fig:[problem23]}
\end{figure}

\begin{figure}[t!]
    \centering
    \captionsetup[sub]{font=scriptsize,labelfont={bf}}
    \begin{subfigure}[b]{0.115\textwidth}
        \includegraphics[width=2cm, height=2.2cm]{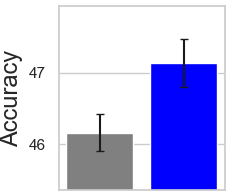}
        
        \caption{ImageNet Cats}
        \label{fig:imagenet-cats}
    \end{subfigure}
    \begin{subfigure}[b]{0.115\textwidth}
        \includegraphics[width=2cm]{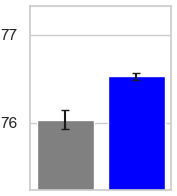}
        \caption{Cifar10}
        \label{fig:problem2-cifar10}
    \end{subfigure}
    \begin{subfigure}[b]{0.115\textwidth}
        \includegraphics[width=2cm]{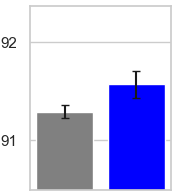}
        \caption{Cifar10 VGG}
        \label{fig:problem4-vgg-cifar10}
    \end{subfigure}
    \begin{subfigure}[b]{0.115\textwidth}
        \includegraphics[width=2cm]{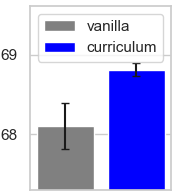}
         \caption{Cifar100 VGG}
        \label{fig:problem4-vgg-cifar100}
    \end{subfigure}
    \vspace*{-2mm}
    \caption{Curriculum by transfer learning. Bars indicate the average final accuracy, and error bars indicate the STE. We performed 25 repetitions in (a), 5 in (b) and 3 in (c,d). (a) Cats subset of ImageNet. (b) CIFAR-10, trained on a small network. (c, d) CIFAR-10 and CIFAR-100 respectively, trained on the VGG network.}
\label{fig:[problem4-vgg-cifar100]}
%\vspace{-0.4cm}
\end{figure}

\paragraph{Cases 2 and 3:}
Similar empirical evaluation as in \textbf{case 1}, using the same moderate size network to classify two benchmark datasets. %In both cases, we optimize using the grid-search mentioned above the learning rate hyper-parameters for both vanilla and curriculum test cases. In addition, we optimize the pacing parameters in the curriculum case. 
The results for CIFAR-10 are shown in Fig.~\ref{fig:problem2-cifar10} and for CIFAR-100 in Fig.~\ref{fig:[problem23]}. Like before, test accuracy with curriculum learning increases faster and reaches better final performance in both cases, as compared to the \emph{vanilla} test condition. The beneficial effect of CL is larger when classifying the more challenging CIFAR-100 dataset. 

\paragraph{Cases 4 and 5:}
Similar empirical evaluation as in \textbf{case 1}, using a competitive public-domain architecture. Specifically, we use the Inception-based \emph{transfer scoring function} to train a VGG-based network \citep{liu2015very} to classify the CIFAR-10 and CIFAR-100 datasets. Differently from the previous cases, here we use the \emph{varied exponential pacing function} with a slightly reduced learning rate, as it has the fewest hyper-parameters to tune since learning rate parameters do not need to be re-tuned, an important factor when training large networks. Results for CIFAR-10 are shown in Fig.~\ref{fig:problem4-vgg-cifar10} and for CIFAR-100 in Fig.~\ref{fig:problem4-vgg-cifar100}; in both cases, no data augmentation has been used. The results show the same qualitative results as in the previous cases; CL gives a smaller benefit, but the benefit is still significant.

\paragraph{Case 6:}
Similar to \textbf{case 1}, using the same moderate size network to distinguish 7 classes of cats from the ImageNet dataset (see \app~\ref{app: imagenet-cats} for details). The results are shown in Fig.~\ref{fig:imagenet-cats}. Again, the test accuracy in the curriculum test condition increases faster and achieves better final performance with curriculum, as compared to the \emph{vanilla} test condition.

% \begin{figure}[ht!]
%     \centering
%     \begin{subfigure}[b]{0.5\textwidth}
%         \includegraphics[width=\textwidth]{graphs/imagenet_cats.png}
% %         \caption{\textbf{Case 2}: CIFAR-10}
% %         \label{fig:problem2-cifar10-curves}
%     \end{subfigure}
%     \caption{Results in \textbf{case 5}, with Inception-based \emph{transfer scoring function} and \emph{fixed exponential pacing function}. Inset: zoom-in on the final iterations, for better visualization. Error bars show STE after 25 repetitions.\label{fig:imagenet-cats}}
% \end{figure}

%\subsection{Self-Taught Curriculum Learning vs. Self-Paced Learning}
\subsection{Curriculum by Bootstrapping}

The \emph{self-taught scoring function} is based on the loss of training points with respect to the final hypothesis of a trained network - the same network architecture pre-trained without a curriculum. Using this scoring function, training is re-started from scratch. Thus defined, \emph{curriculum by bootstrapping} may seem closely related to the idea of \emph{Self-Paced Learning} (SPL), an iterative procedure where higher weights are given to training examples that have lower cost with respect to the current hypothesis. %Given the relevant recent results by \citet{weinshallcurriculum} as discussed in the introduction, twe define the notion of ``objectively easy" examples and ``subjectively easy" examples -- objectively easy examples are easy because of some inherent property they possess and should be easy for many networks, while subjectively easy examples are easy only to a specific network, for any number of reasons (for example, its very similar to an example in its train set). The results of \citep{weinshallcurriculum}, suggest that using the self-paced \emph{scoring function} might show the network mostly subjectively easy examples, which can impede the network training procedure, as it will over-fit on these subjectively easy examples. 
There is, however, a very important difference between the methods: SPL determines the \emph{scoring function} based on the loss with respect to the current hypothesis (or network), while \emph{bootstrapping CL} scores each point by its loss with respect to the target hypothesis. %Note that when using CL to optimize the linear regression loss (see introduction), \emph{self-taught curriculum} and self-paced learning are discordant. 
SPL, it appears, has not yet been introduced to deep neural networks in a way that  benefits accuracy. 

To compare the \emph{self-taught scoring function} and the \emph{self-paced scoring function}, we investigate their effect on CL in the context of test \textbf{case 1}. Final accuracy results are shown in Fig.~\ref{fig:problem1-summary} (the entire learning curves are depicted in \app~\ref{app: addition-details}, Fig.~\ref{app: fig:self-pace}). As expected, we see that \emph{bootstrapping CL} improves test accuracy throughout the entire learning session. On the other hand, CL training using the \emph{self-paced scoring function} decreases the test accuracy throughout. This decrease is more prominent at the beginning of the learning, where most of the beneficial effects of the curriculum are observed, suggesting that the \emph{self-paced scoring function} can significantly delay learning.  

%We note that, at least in principle, the \emph{self-taught scoring function} can be used repeatedly: after training the network using a curriculum, we can use its confidence score to define a new \emph{scoring function} and retrain the network once again from scratch. However, in practice we did not observe any benefit to subsequent trainings.

\subsection{Alternative Pacing Functions}

\begin{figure}[t!]
    \centering
    \begin{subfigure}[b]{0.49\textwidth}
        \includegraphics[width=\textwidth]{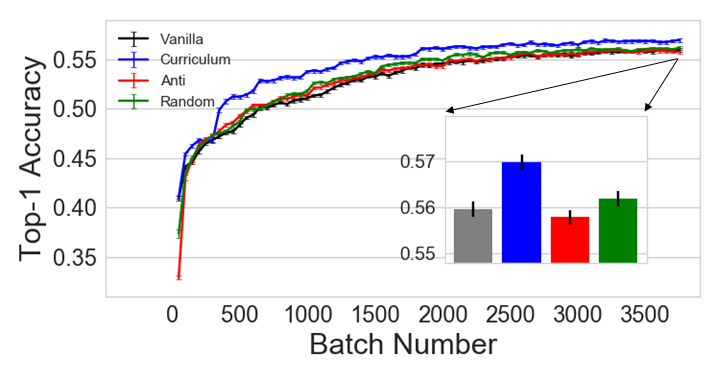}
        % \caption{\emph{Single step pacing function}}
        %  \label{fig:problem1-single-step-curves}
    \end{subfigure}
    \vspace{-0.8cm}
%     \begin{subfigure}[b]{0.49\textwidth}
%         \includegraphics[width=\textwidth]{graphs/problem1_fixed_vs_varied.png}
%         \caption{Varied vs. fixed exponential pacing.}
%         \label{fig:problem1-varied-vs-fixed}
%     \end{subfigure}
%     \vspace{-0.5cm}
    ~ %add desired spacing between images, e. g. ~, \quad, \qquad, \hfill etc. 
    
    \caption{Results in \textbf{case 1}, with Inception-based \emph{transfer scoring function} and \emph{Single step pacing function}. Inset: bars indicating the average final accuracy in each condition over the last few iterations. Error bars indicate the STE after 50 repetitions.}\label{fig:problem1-single-step-curves}
\vspace{-0.5cm}
\end{figure}

\paragraph{Single step pacing.}
Curriculum learning can be costly, and it affects the entire learning protocol via the \emph{pacing function}. At the same time, we observed empirically that the main effect of the procedure seems to have taken place at the beginning of training. %, in agreement with the theory \citep{weinshallcurriculum}. 
This may be due, in part, to the fact that the proposed \emph{scoring function} $f$ is based on transfer from another network trained on a different dataset, which only approximates the unknown \emph{ideal scoring function}. Possibly, since the \emph{scoring function} is based on one local minimum in a complex optimization landscape which contains many local minima, the score given by $f$ is more reliable for low scoring (easy) examples than high scoring (difficult) examples, which may be in the vicinity of a different local minimum.  

% \begin{figure}[ht!]
%     \centering
% %     \begin{subfigure}[b]{0.495\textwidth}
% %         \includegraphics[width=\textwidth]{graphs/problem1_single_step_curriculum_vanilla_anti_random.png}
% %         \caption{\emph{Single step pacing function}}
% %          \label{fig:problem1-single-step-curves}
% %     \end{subfigure}
%     \begin{subfigure}[b]{0.49\textwidth}
%         \includegraphics[width=\textwidth]{graphs/problem1_fixed_vs_varied.png}
%         \caption{Varied vs. fixed exponential pacing.}
%         \label{fig:problem1-varied-vs-fixed}
%     \end{subfigure}
%     \vspace{-0.5cm}
%     ~ %add desired spacing between images, e. g. ~, \quad, \qquad, \hfill etc. 
%       %(or a blank line to force the subfigure onto a new line)
%     \caption{Results in \textbf{case 1}, with Inception-based \emph{transfer scoring function}. (a) \emph{Single step pacing function}, (b) \emph{varied Exponential pacing  function}. Inset: bars indicating the average final accuracy in each condition over the last few iterations. Error bars indicate the STE after 50 repetitions. \guy{fix caption}}\label{fig:two-jumps1}
% \end{figure}

Once again we evaluate \textbf{case 1}, using the \emph{transfer scoring function} and the \emph{single step pacing function}.
% We tune the pacing and learning rate hyper-parameters for all test conditions. All tunings were done by simple grid-search, confirmed by cross-validation\guy{didn't we write this in the methodology?}.
We see improvement in test accuracy in the curriculum test condition, which resembles the improvement achieved using the \emph{exponential pacing}. Results are shown in Fig.~\ref{fig:problem1-single-step-curves}. It is important to note that this \emph{pacing function} ignores most of the prior knowledge provided by the \emph{scoring function}, as it only uses a small percent of the easiest examples, and yet it achieves competitive results. Seemingly, in our empirical setup, most of the power of CL lies at the beginning of training.

\subsection{Analysis of Scoring Function}%: Analysis and Empirical Evaluation}

\begin{figure}[ht!]
    \centering
	\begin{subfigure}[b]{0.3\textwidth}
        \includegraphics[width=\textwidth]{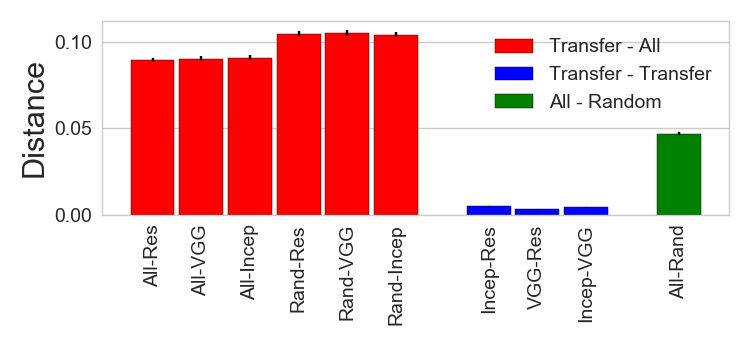}
         \caption{Distances between mean gradients.}
        \label{fig:problem-1-mean-gradient}
    \end{subfigure}
    ~ %add desired spacing between images, e. g. ~, \quad, \qquad, \hfill etc. 
	\hfill
    \begin{subfigure}[b]{0.15\textwidth}
        \includegraphics[width=\textwidth]{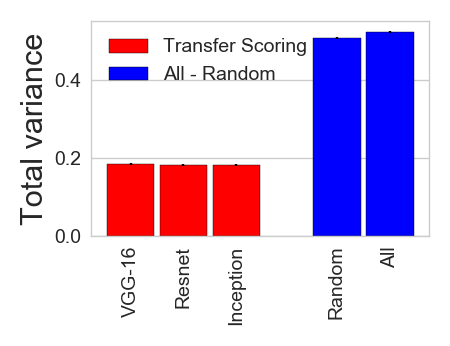}
        \caption{Total variance.}
        \label{fig:problem-1-mean-std}
    \end{subfigure}
    \vspace*{-3mm}
    \caption{(a) Distance between the mean gradient direction of preferred examples under different \emph{scoring functions}. Each bar corresponds to a pair of mean gradients in two different conditions, %(from left to right): random-ResNet, random-VGG, entire data-Inception, VGG-ResNet, Inception-ResNet, entire data-Random, entire data-VGG, Inception-VGG, entire data-ResNet and random-Incetion.  
see text. (b) The total variance of each set of gradients.}
\label{fig:problem-1-gradient-std}
%\vspace{-0.2cm}
\end{figure}

In order to analyze the effects of transfer based \emph{scoring functions}, we turn to analyze the gradients of the network's weights w.r.t the empirical loss. We evaluate the gradients using a pre-trained \emph{vanilla} network in the context of \textbf{case 1}.  First, for each method and each \emph{scoring function}, we collect the subset of points used to sample the first mini-batch according to the \emph{pacing function} $g_{\vartheta}(1)$\footnote{In this experiment $g_{\vartheta}(1)$ is set such that it corresponds to $10\%$ of the data or 250 examples. This number was set arbitrarily, with similar qualitative results obtained for \comment{a large range of} other choices.}. For comparison, we also consider the set of all training points, which are used to compute the exact gradient of the empirical loss in batch learning using GD. We then compute the corresponding set of gradients for the training points in each of these subsets of training points, treating each layer's parameters as a single vector, and subsequently estimate the gradients' mean and total variance\footnote{%As customary, total 
Total variance denotes the trace of the covariance matrix.}. We use these measurements to evaluate the coherence of the gradients in the first mini-batch of each \emph{scoring function}. The Euclidean distance between the mean gradient in the different conditions is used to estimate the similarity between the different \emph{scoring functions}, based on the average preferred gradient. 
We compare the set of gradients defined by using three \emph{transfer scoring functions}, which differ in the teacher network used for scoring the points: 'VGG-16', 'ResNet', and 'Inception'.  We include in the comparison the gradients of the \emph{random scoring function} denoted 'Random', and the gradients of the whole batch of training data denoted 'All'. Results are shown in Fig.~\ref{fig:problem-1-gradient-std}. 

We see in Fig.~\ref{fig:problem-1-mean-gradient} - blue bars - that the average gradient vectors, computed based on the 3 \emph{transfer scoring functions}, are quite similar to each other. This suggests that they are pointing towards nearby local minima in parameters space. We also see - green bar - that the average gradient vector computed using a random subset of examples resembles the exact empirical gradient computed using all the training data. This suggests that a random subset provides a reasonable estimate of the true empirical gradient. The picture changes completely when we compute  - red bars - the distance between the average gradient corresponding to one of the 3 transfer \emph{scoring functions}, and the average random gradient or the empirical gradient. The large distances suggest that CL by transfer stirs the weights towards different local minima in parameter space as compared to \emph{vanilla} training.  

We see in Fig.~\ref{fig:problem-1-mean-std} that the total variance for the 3 transfer \emph{scoring functions} is much smaller than the total variance of some random subset of the whole training set. This intuitive result demonstrates the difference between training with easier examples and training with random examples, and may -- at least partially -- explain the need for a different learning rate when training with easier examples.  

\subsection{Summary of Results}

%\vspace{-0.2cm}
\begin{figure}[ht!]
% \vspace{-0.4cm}
%     \centering
%     \begin{subfigure}[b]{0.585\textwidth}
%         \includegraphics[width=\textwidth]{graphs/problem1_summary.png}
%         \vspace{-0.64cm}
%          \caption{}%Learning curves and final performance.}
% 	\label{fig:problem1-summary-left}
%     \end{subfigure}
%     ~ %add desired spacing between images, e. g. ~, \quad, \qquad, \hfill etc. 
%       %(or a blank line to force the subfigure onto a new line)
    \begin{subfigure}[b]{0.49\textwidth}
    \includegraphics[height=4cm, width=\textwidth]{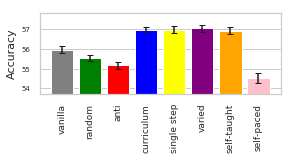}
%         \vspace{-1.9cm}
%          \caption{}%Area under the curve for each method.}
% 	\label{fig:problem1-summary-right}
    \end{subfigure}
    \vspace*{-8mm}
    \caption{Results in \textbf{case 1}, bars showing final accuracy in percent for all test conditions. Error bars indicate STE after 50 repetitions.}
    \label{fig:problem1-summary}
%\label{fig:problem1-summary-curves}
\end{figure}

Fig.~\ref{fig:problem1-summary} summarizes the main results presented in this section, including: curriculum with an Inception-based \emph{scoring function} for (i) \textsl{fixed} exponential pacing (denoted \textsl{curriculum}), (ii) \textsl{single step} pacing, and (iii) \textsl{varied} exponential pacing. It also shows curriculum with fixed exponential pacing for (iv) \textsl{self-taught} scoring, and (v) \textsl{self-paced} scoring. In addition, we plot the control conditions of \textsl{vanilla}, \textsl{anti}-curriculum, and \textsl{random}. The bars depict the final accuracy in each condition. %, and error bars that represent the standard error after 50 repetitions. 
All the curriculum conditions seem to improve the learning accuracy throughout the entire learning session while  converging to similar performance, excluding the self-paced \emph{scoring function} which impairs learning. While different conditions seem to improve the final accuracy in a similar way, the results of \textsl{curriculum by transfer} are easier to obtain, and are more robust (see \app~\ref{app: robustness} for details).

\section{Theoretical Analysis}
\label{sec:theory}

Let $\gH$ denote a set of hypotheses $h_\vartheta$ defined by the vector of hyper-parameters $\vartheta$. Let $L_\vartheta(\bx_i)$ denote the loss of hypothesis $h_\vartheta$ when given example $\bx_i$. In order to compute the best hypothesis $h_{\tilde\vartheta}$ from the data, one commonly uses the Empirical Risk Minimization (ERM) framework where\footnote{$\hat A$, for any operator $A$, denotes the empirical estimate of $A$.} 
\begin{equation}
\begin{split}
\label{eq:ERM}
    \Ls(\vartheta)&=\hat\E[L_\vartheta]=\frac{1}{N}\sum_{i=1}^N L_\vartheta(\bx_i) \\ 
    \tilde\vartheta &= \argmin_\vartheta \Ls(\vartheta)
\end{split}
\end{equation}
$\Ls(\vartheta)$ denotes the empirical loss given the observed data, thus defining the Risk of choosing hypothesis $h_\vartheta$. (\ref{eq:ERM}) can be rewritten as follows:
\begin{equation*}
%\label{eq:ML}
\begin{split}
    \tilde\vartheta &= \argmin_\vartheta \sum_{i=1}^N L_\vartheta(\bx_i) = \argmax_\vartheta \exp(-\sum_{i=1}^N L_\vartheta(\bx_i)) \\ &= \argmax_\vartheta \prod_{i=1}^N e^{-L_\vartheta(\bx_i)} \triangleq \argmax_\vartheta \prod_{i=1}^N \alpha P(\vartheta\vert_{\displaystyle{\bx_i}})
\end{split}
\end{equation*}
Thus ERM can be interpreted as Maximum Likelihood (ML) estimation with probability defined by the loss function as $P(\vartheta\vert_{\displaystyle{\bx}}) \propto e^{-L_\vartheta(\bx)}$.

The choice of loss $L_\vartheta(\bx)$, and the choice of the estimation framework used to select some optimal hypothesis $h_{\tilde\vartheta}$, are somewhat arbitrary. In a similar manner we may choose to maximize the average \emph{Utility} $U_\vartheta(\bx)= e^{-L_\vartheta(\bx)}$ of the observed data, %. Based on the equivalence to ML pointed out above, we may define Utility by the expected \emph{Value} (Value being proportional to likelihood) of the observed data 
which is defined as follows
\begin{equation}
\begin{split}
\label{eq:MU}
    \gU(\vartheta)&=\hat\E[U_\vartheta]=\frac{1}{N}\sum_{i=1}^N U_\vartheta(\bx_i)\triangleq\frac{1}{N}\sum_{i=1}^N e^{-L_\vartheta(\bx_i)} \\
    \tilde\vartheta &= \argmax_\vartheta \gU(\vartheta)
\end{split}
\end{equation}
The ERM formulation defined in (\ref{eq:ERM}) is different from the empirical utility maximization formulation defined in (\ref{eq:MU}). Both formulations can be similarly justified from first principles.

The \emph{scoring function} in curriculum learning effectively provides a Bayesian prior for data sampling. This can be formalized as follows:
% \begin{equation}
% \begin{split}
\begin{align}
\label{eq:bayesian-utility}
    \gV(\vartheta)&=\hat\E_\vp[U_\vartheta]=\sum_{i=1}^N U_\vartheta(\bx_i)p(\bx_i)=\sum_{i=1}^N e^{-L_\vartheta(\bx_i)}p_i \notag\\
    \tilde\vartheta &= \argmax_\vartheta \gV(\vartheta)
\end{align}
% \end{split}
% \end{equation}
Above $p_i=p(\bx_i)$ denotes the induced prior probability, which is determined by the \emph{scoring function} and \emph{pacing function} of the curriculum algorithm. Thus $p(\bx_i)$ will always be a non-increasing function of the difficulty level of $\bx_i$. In our algorithm, $p(\bx_i)=\frac{1}{M}$ for $M$ training points whose difficulty score is below a certain threshold, and $p(\bx_i)=0$ otherwise. The threshold is determined by the \emph{pacing function} which drives a monotonic increase in the number of points $M$, thus changing the optimization function in a corresponding manner.

From (\ref{eq:bayesian-utility}), $\gV(\vartheta)$ is a function of $\vartheta$ which is determined by the correlation between two random variables, $U_\vartheta(X)$ and $p(X)$. We rewrite (\ref{eq:bayesian-utility}) as follows
\begin{equation}
\label{eq:cov}
\begin{split}
    \gV(\vartheta)&=\sum_{i=1}^N (U_\vartheta(\bx_i)-\hat\E[U_\vartheta])(p_i-\hat\E[p])+N\hspace{1pt}\hat\E[U_\vartheta]\hat\E[p]\\ & = \hat\Cov[U_\vartheta,p]+N\hspace{1pt}\hat\E[U_\vartheta]\hat\E[p]=\gU(\vartheta) + \hat\Cov[U_\vartheta,p]
\end{split}
\end{equation}
% Since $\hat\E[p]=\frac{1}{N}\sum_{i=1}^N p(X_i)=\frac{1}{N}$ while $\gU(\vartheta)=\hat\E[U_\vartheta]$, we can write
% \begin{equation}
% \label{eq:cov}
%     \gV(\vartheta)= \gU(\vartheta) + \hat\Cov[U_\vartheta,p]
% \end{equation}
This proves the following result:
\begin{proposition}
The difference between the expected utility when computed with and without prior $\vp$ is the covariance between the two random variables $U_\vartheta(X)$ and $p(X)$.
\end{proposition}

Curriculum learning changes the landscape of the optimization function over the hyper-parameters $\vartheta$ from $\gU(\vartheta)$ to $\gV(\vartheta)$. Intuitively, (\ref{eq:cov}) suggests that if the induced prior probability $\vp$, which defines a random variable over the input space $p(X)$, is positively correlated with the optimal utility $U_{\tilde\vartheta} (X)$, and more so than with any other $U_{\vartheta} (X)$, then the gradients in the direction of the optimal parameter $\tilde\vartheta$ in the new optimization landscape may be overall steeper. 

%\begin{equation*}
%    p(\bx_i) = P(\bx_i \vert_{\displaystyle{\tilde\vartheta}}) \propto %P(\tilde\vartheta\vert_{\displaystyle{\bx_i}})
%\end{equation*}

% In order to state the next theorem we make the following assumptions:
% \begin{defn}[Assumptions]
% \label{def:assumptions}
% $ $\\
% \vspace{-4mm}
% \begin{enumerate}
% \item
% $\argmax_\vartheta \gU(\vartheta) = \argmax_\vartheta \hat\Cov[U_\vartheta,p] = \tilde\vartheta$
% \item
% $\Var[p(X)] \le 1$
% \end{enumerate}
% \end{defn}
% The first assumption implies that the 

% \begin{theorem}
% \label{thm:1}
More precisely, assume that $\tilde\vartheta$ maximizes the covariance between $p_\vartheta (X)$ and the utility $U_\vartheta (X)$, namely
\begin{equation}
\label{eq:assumption}
\argmax_\vartheta \gU(\vartheta) = \argmax_\vartheta \hat\Cov[U_\vartheta,p] = \tilde\vartheta
\end{equation}
\begin{proposition}
\label{proposition:1}
For any curriculum %whose induced probability $p(X)$ satisfies
satisfying (\ref{eq:assumption}):
\begin{enumerate}
\item
$\tilde\vartheta = \argmax_\vartheta \gU(\vartheta) = \argmax_\vartheta \gV(\vartheta)$
\item
$\gV(\tilde\vartheta)-\gV(\vartheta)\ge\gU(\tilde\vartheta)-\gU(\vartheta)~~~~\forall\vartheta$
\end{enumerate}
\end{proposition}
% \end{theorem}
% \begin{proof}
% Claim 1 follows directly from (\ref{eq:assumption}), while for claim 2:
% \begin{equation*}
% \begin{split}
% \gV(\tilde\vartheta)&-\gV(\vartheta)=\gV(\tilde\vartheta)-\gU(\vartheta) - \hat\Cov[U_\vartheta,p] \\
% &\geq \gV(\tilde\vartheta)-\gU(\vartheta) - \hat\Cov[U_{\tilde{\vartheta}},p] = \gU(\tilde\vartheta)-\gU(\vartheta)
% \end{split}
% \end{equation*}
% \end{proof}

Proof can be found in \app~\ref{app: theoretical}. We conclude that when assumption (\ref{eq:assumption}) holds, the modified optimization landscape induced by curriculum learning has the same global optimum $\tilde\vartheta$ as the original problem. In addition, the modified optimization function in the parameter space $\vartheta$ has the property that the global maximum at $\tilde\vartheta$ is more pronounced. %In some cases, this implies that local differences are biased towards the optimal solution, as compared to the original optimization problem. This bias might partially explain why we see faster convergence in curriculum learning. \guy{too many ``partially"/``could imply"/ ``may"/ ``in some cases". re-wrote to remove some of those, i hope i don't state something too precisely}

We define an \emph{ideal curriculum} to be the prior corresponding to the optimal hypothesis (or one of them, if not unique): 

\begin{equation*}
p_i=\frac{e^{-L_{\tilde\vartheta}(\bx_i)}}{C}, ~~~ C=\sum_{i=1}^N e^{-L_{\tilde\vartheta}(\bx_i)}
\end{equation*}
%For simplicity of notation, assume that $\tilde\vartheta = \argmax_\vartheta \gU(\vartheta)$ is unique \guy{why do we care if its unique?}. 
From\footnote{Henceforth we will assume that $N\rightarrow\infty$, so that the estimation symbol $\hat a$ can be omitted.} (\ref{eq:cov}):
\begin{equation*}
    \gV(\vartheta)=\gU(\vartheta) + \frac{1}{C}\Cov[U_\vartheta,U_{\tilde\vartheta}]
\end{equation*}

The utility at the optimal point $\tilde\vartheta$ in parameter space is:
\begin{equation}
\label{eq:vthe}
    \gV(\tilde\vartheta)=\gU(\tilde\vartheta) + \frac{1}{C}\Cov[U_{\tilde\vartheta},U_{\tilde\vartheta}] = \gU(\tilde\vartheta) + \frac{1}{C}\Var[U_{\tilde\vartheta}]
\end{equation}
In any other point
\begin{equation}
\begin{split}
\label{eq:corr-prior}
    \gV(\vartheta)&=\gU(\vartheta) + \frac{1}{C}\Cov[U_\vartheta,U_{\tilde\vartheta}] \\
    &\le \gU(\tilde\vartheta) + \frac{1}{C}\sqrt{\Var[U_\vartheta]\Var[U_{\tilde\vartheta}]}
\end{split}
\end{equation}
Note that if $\Var[U_\vartheta]=b~\forall\vartheta$ for some constant $b$, then assumption (\ref{eq:assumption}) immediately follows from (\ref{eq:corr-prior}): 
\begin{equation*}
    \gV(\vartheta)\le\gU(\tilde\vartheta) + \frac{1}{C}\sqrt{b^2} = \gV(\tilde\vartheta) ~\implies~\tilde\vartheta = \argmax_\vartheta \gV(\vartheta)
\end{equation*}
Therefore
\begin{corollary}
When using the ideal curriculum,  Proposition~\ref{proposition:1} holds if the variance of the utility function is roughly constant in the relevant range of plausible parameter values. 
\end{corollary}

From (\ref{eq:vthe}) and (\ref{eq:corr-prior}) we can also conclude the following
\begin{proposition}
\label{proposition:2}
When using the ideal curriculum
\begin{equation*}
\gV(\tilde\vartheta)-\gV(\vartheta)\ge\gU(\tilde\vartheta)-\gU(\vartheta)~~~~\forall\vartheta :\Cov[U_\vartheta,U_{\tilde\vartheta}] \le \Var[U_{\tilde\vartheta}]
\end{equation*}
\end{proposition}
This implies that the optimization landscape is modified to amplify the difference between the optimal parameters vector and all other parameter values whose covariance with the optimal solution (the covariance is measured between the induced prior vectors) is small, and specifically smaller than the variance of the optimum. In particular, this includes all parameters vectors which are uncorrelated (or negatively correlated) with the selected optimal parameter vector.

\paragraph{Discussion:}
Training a network based on its current hypothesis $\vartheta_t$ can be done in one of 2 ways: (i) using a prior which is monotonically increasing with the current utility, as suggested by self-paced learning; (ii) using a prior monotonically decreasing with the current utility, as suggested by hard data mining or boosting. Our analysis suggests that as long as the curriculum is positively correlated with the optimal utility, it can improve the learning; hence both strategies can be effective in different settings. It may even be possible to find a curriculum which is directly correlated with the optimal utility, and that outperforms both methods.

\section*{Acknowledgements}

This work was supported in part by a grant from the Israel Science Foundation (ISF), MAFAT Center for Deep Learning, and the Gatsby Charitable Foundations.

\appendix
\section*{Appendix}
\section{Additional Empirical Results}

\begin{figure}[ht!]
    \centering
    \begin{subfigure}[t]{0.49\textwidth}
        \includegraphics[width=\textwidth]{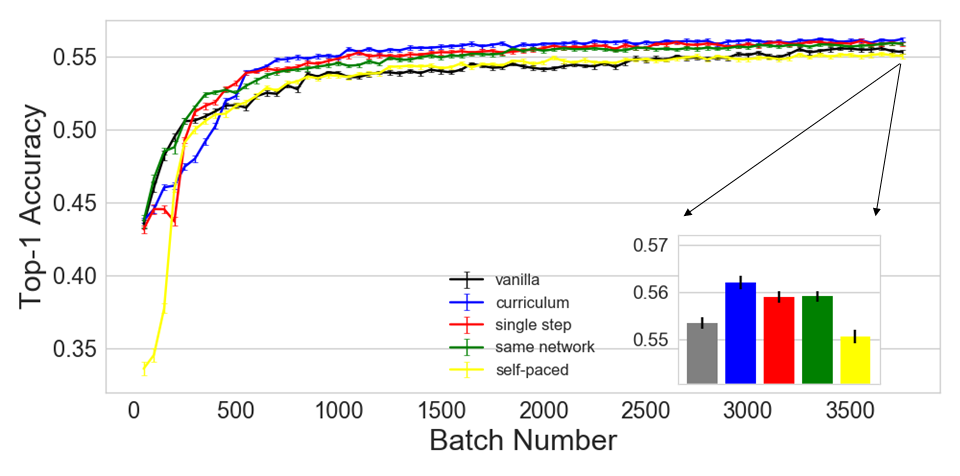}
        %\subcaption{Our results}
        %\label{fig:problem1-ours}
    \end{subfigure}
    \caption{Results under the same conditions as in Fig.~\ref{fig:problem1-summary}, using instead the ``aquatic mammals" CIFAR-100 super-class. Error bars show STE after 50 iterations.}
    \label{fig:appendix-dataset0}
\end{figure}

\paragraph{CL with other CIFAR-100 super-classes.}
\label{app:curriculum-subset-0}
In Section~\ref{sec:empirical} we present results when learning to discriminate the ``small mammals" super-class of CIFAR-100. Similar results can be obtained for other super-classes of CIFAR-100. Each super-class contains 3000 images, divided into 5 related classes of CIFAR-100. Each class contains 600 images divided into 500 train images and 100 test images. Specifically, we tested our method on the super-classes of ``people", ``insects" and ``aquatic mammals" and found that CL trained on these different super-classes shows the same qualitative results. % - both when using \emph{transfer scoring function} or \emph{self-taught scoring function} with either the \emph{single-step pacing function} or the \emph{fixed-exponential pacing function}. We note again, 
We note once again that CL is more effective  in the harder tasks, namely, the super-classes containing classes that are harder to discriminate (measured by lower \emph{vanilla} accuracy). As an example,  Fig.~\ref{fig:appendix-dataset0} shows results using the ``aquatic mammals" super-class, which greatly resembles the results we've seen when discriminating the ``small mammals" super-class (cf. Fig.\ref{fig:problem1-summary}). %Although not tested empirically, we hypothesize that similar results can be obtained on the rest of the CIFAR-100 super-classes using the same framework. 

\paragraph{Transfer based scoring function.}
\label{appendix:transfer_networks_comparison}

In the experiments described in Section~\ref{sec:empirical}, when using the \emph{transfer scoring function} defined in Section~\ref{sec:scoring-pacing-def}, we use the pre-trained Inception network available from https://github.com/Hvass-Labs/TensorFlow-Tutorials. We normalized the data similarly to the normalization done for the neural network, resized it to $299\times 299$, and ran it through the Inception network. We then used the penultimate layer's activations as features for each training image, resulting in 2048 features per image. Using these features, we trained a Radial Basis Kernel (RBF) SVM \citep{scholkopf1997comparing} and used its confidence score to determine the difficulty of each image. The confidence score of the SVM was provided by \emph{sklearn.svm.libsvm.predict\_proba} from Python's Sklearn library and is based on cross-validation.

\begin{figure}[ht!]
    \centering
    \begin{subfigure}[t]{0.475\textwidth}
        \includegraphics[width=\textwidth]{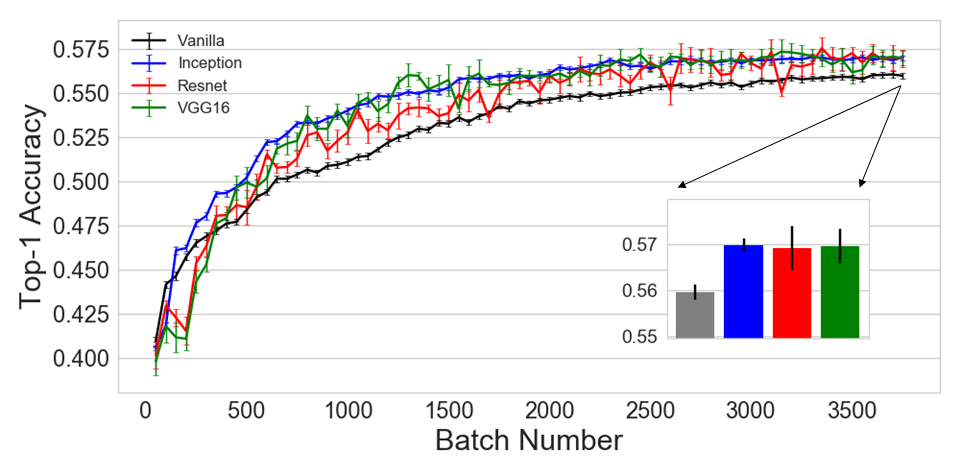}
        \subcaption{Three competitive networks trained on Imagenet.}
        \label{fig:appendix-vgg-resnet}
    \end{subfigure}
    \begin{subfigure}[t]{0.475\textwidth}
        \includegraphics[width=\textwidth]{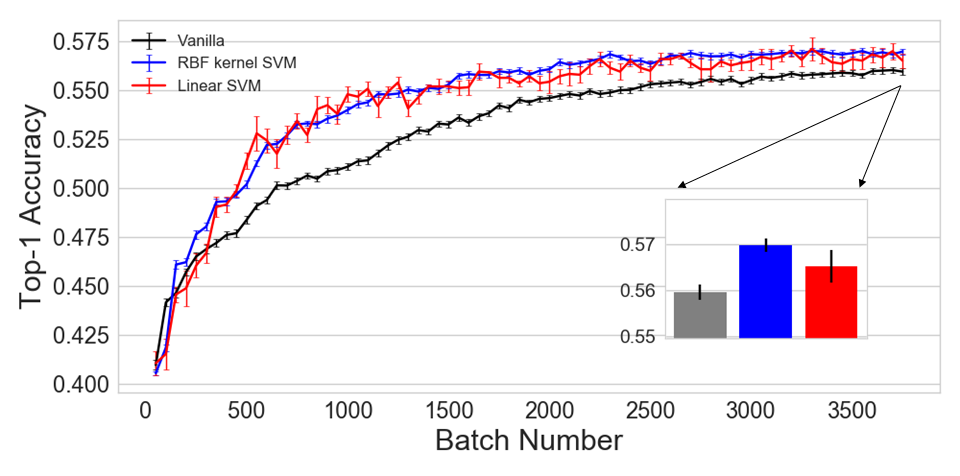}
        \subcaption{Two different classifiers.}
        \label{fig:rbf_linear}
    \end{subfigure}
    \caption{Results in \textbf{case 1}. Comparing different variants of the \emph{transfer scoring function}. The inset bars show the final accuracy of the learning curves. The error bars shows STE after 50 repetitions for the \emph{vanilla} and Inception conditions with \emph{RBF kernel SVM}, and 5 repetitions for the \emph{ResNet}, \emph{VGG-16} and the \emph{Linear SVM} conditions. (a) Comparing different teacher networks. (b) Comparing different classifiers for the hardness score.}
    \label{}
\end{figure}

Choosing Inception as the teacher and RBF SVM as the classifier was a reasonable arbitrary choice --  the same qualitative results are obtained when using other large networks trained on ImageNet as teachers, and other classifiers to establish a confidence score. Specifically, we repeated the experiments with a \emph{transfer scoring function} based on the pre-trained VGG-16 and ResNet networks, which are also trained on Imagenet. The curriculum method using the \emph{transfer scoring function} and \emph{fixed exponential pacing function} are shown in Fig.~\ref{fig:appendix-vgg-resnet}, demonstrating the same qualitative results. Similarly, we used a linear SVM instead of the RBF kernel SVM with similar results, as shown in Fig.~\ref{fig:rbf_linear}. We note that the STE error bars are relatively large for the control conditions described above because we only repeated these conditions 5 times each, instead of 50 as in the main experiments. 

\paragraph{Varied exponential pacing.}
\label{app: Varied_exponnetial}
%We saw that the fixed exponential pacing function used in \citet{weinshallcurriculum} and generalized above, can be used to improve the final accuracy of a network when we optimize its learning rate separately. Separate optimization to the learning rate is needed due to an indirect manipulation which is done by the pacing function. 
We define \emph{Varied exponential pacing} similarly to \emph{fixed exponential pacing}, only allowing to change the \textsl{step\_length} for each step. Theoretically, this method results in additional hyper-parameters equal to the number of performed steps. In practice, to avoid an unfeasible need to tune too many hyper-parameters, we vary only the first two \textsl{step\_length} instances and fix the rest. This is reasonable as most of the power of the curriculum lies in the first few \textsl{steps}. Formally, \emph{Varied exponential pacing} is given by:
\begin{equation*}
\begin{split}
g_{\vartheta}(i) = &\min\left(starting\_percent\cdot increase^{z(i)} ,1\right)\cdot N\\
&~~z(i) = {\sum_{k=1}^{\#steps}\1_{\left[i>step\_length_k\right]}}
\end{split}
\end{equation*}
where \textsl{starting\_percent} and \emph{increase} are the same as \emph{fixed exponential pacing}, while \textsl{step\_length} may vary in each step. 
The total number of \textsl{steps} can be calculated from \textsl{starting\_percent} and \emph{increase}:
\begin{equation*}
\#step=\ceil{-\log_{increase}(starting\_percent)}
\end{equation*}

This \emph{pacing function} allows us to run a CL procedure without the need for further tuning of learning rate. The additional parameters added by this method control directly the number of epochs the network trains on each dataset size. If tuned correctly, this allows the pacing function to mitigate most of the indirect effect on the learning rate, as it can choose fewer epochs for data sizes which has a large effective learning rate.

\begin{figure}[ht!]
    \centering
%     \begin{subfigure}[b]{0.495\textwidth}
%         \includegraphics[width=\textwidth]{graphs/problem1_single_step_curriculum_vanilla_anti_random.png}
%         \caption{\emph{Single step pacing function}}
%          \label{fig:problem1-single-step-curves}
%     \end{subfigure}
    \begin{subfigure}[b]{0.49\textwidth}
        \includegraphics[width=\textwidth]{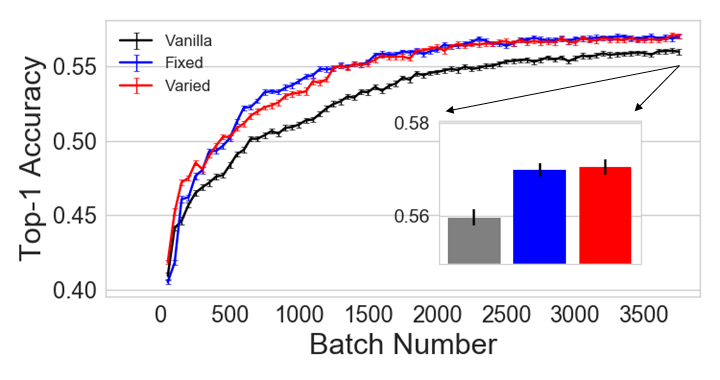}
    \end{subfigure}
    \vspace{-0.5cm}
    ~ %add desired spacing between images, e. g. ~, \quad, \qquad, \hfill etc. 
    \caption{Comparing \emph{fixed exponential pacing} to \emph{varied exponential pacing} in \textbf{case 1}, with Inception-based \emph{transfer scoring function}. Inset: bars indicating the average final accuracy in each condition over the last few iterations. Error bars indicate the STE after 50 repetitions.}\label{fig:two-jumps1}
\end{figure}

Once again we evaluate \textbf{case 1}, fixing the learning rate parameters to be the same as in the \emph{vanilla} test condition, while tuning the remaining hyper-parameters as described in Section~\ref{sec:scoring-pacing-def} using a grid search with cross-validation. We see improvement in the accuracy throughout the entire learning session, although smaller than the one observed with \emph{fixed exponential pacing}. However, decreasing the learning rate of the \emph{vanilla} by a small fraction and then tuning the curriculum parameters achieves results which are very similar to the \emph{fixed exponential pacing}, suggesting that this method can almost completely nullify the indirect manipulation of the learning rate in the \emph{fixed exponential pacing} function. These results are shown in Fig.~\ref{fig:two-jumps1}.

\section{Extended Discussion}
%\vspace{-0.2cm}
\begin{figure}[ht!]
% \vspace{-0.4cm}
%     \centering
%     \begin{subfigure}[b]{0.585\textwidth}
%         \includegraphics[width=\textwidth]{graphs/problem1_summary.png}
%         \vspace{-0.64cm}
%          \caption{}%Learning curves and final performance.}
% 	\label{fig:problem1-summary-left}
%     \end{subfigure}
%     ~ %add desired spacing between images, e. g. ~, \quad, \qquad, \hfill etc. 
%       %(or a blank line to force the subfigure onto a new line)
    \begin{subfigure}[b]{0.49\textwidth}
    \includegraphics[height=4cm, width=\textwidth]{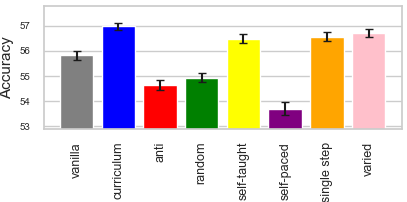}
%         \vspace{-1.9cm}
%          \caption{}%Area under the curve for each method.}
% 	\label{fig:problem1-summary-right}
    \end{subfigure}
    \caption{Results in \textbf{case 1}, when using the AUC as the grid-search optimization criteria. Bars showing final accuracy in percent for all test conditions. Error bars indicate STE after 50 repetitions.}
    \label{app: fig:problem1-summary-trapz}
%\label{fig:problem1-summary-curves}
\end{figure}

%\vspace{-0.2cm}
\begin{figure}[ht!]
% \vspace{-0.4cm}
%     \centering
%     \begin{subfigure}[b]{0.585\textwidth}
%         \includegraphics[width=\textwidth]{graphs/problem1_summary.png}
%         \vspace{-0.64cm}
%          \caption{}%Learning curves and final performance.}
% 	\label{fig:problem1-summary-left}
%     \end{subfigure}
%     ~ %add desired spacing between images, e. g. ~, \quad, \qquad, \hfill etc. 
%       %(or a blank line to force the subfigure onto a new line)
    \begin{subfigure}[b]{0.49\textwidth}
    \includegraphics[height=4cm, width=\textwidth]{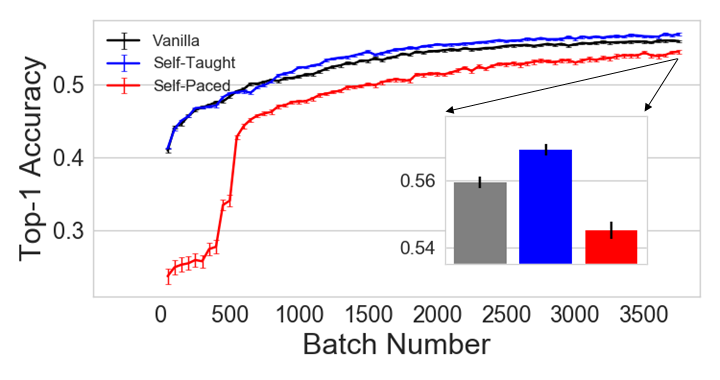}
%         \vspace{-1.9cm}
%          \caption{}%Area under the curve for each method.}
% 	\label{fig:problem1-summary-right}
    \end{subfigure}
    \caption{Self-taught learning vs. self-paced learning. Results are in \textbf{case 1} with the Inception-based \emph{transfer scoring function}. Inset: bars indicating the average final accuracy in each condition, over the last few iterations. Error bars indicate the STE after 50 repetitions}
    \label{app: fig:self-pace}
%\label{fig:problem1-summary-curves}
\end{figure}
\paragraph{Self-taught bootstrapping}
\label{app:self-bootsrapping}
In principle, the \emph{self-taught scoring function} can be used repeatedly to boost the performance of the network indefinitely: after training the network using a curriculum, we can use its confidence score to define a new \emph{scoring function} and retrain the network from scratch. However, \emph{scoring functions} created by repeating this procedure tend to accumulate errors: once an example is misclassified as being easy, this example will be shown more often in subsequent iterations, making it more likely to be considered easy. In practice, we did not observe any benefit to repeated bootstrapping, and even observed an impairment after a large number of repetitions.

\subsection*{Fair comparison in parameter tuning}
\label{app:fair}

When using the moderate size hand-crafted network (\textbf{cases 1, 2, 3} and \textbf{6}), learning rate tuning is done for the \emph{vanilla} case as well. In these cases, for the \emph{curriculum}, \emph{anti-curriculum} and \emph{random} test conditions, we perform a coarse grid search for the \emph{pacing} hyper-parameters as well as the learning rate hyper-parameters, with an identical range of values for all conditions. For the \emph{vanilla} condition, there are no \emph{pacing} hyper-parameters. Therefore, we expand and refine the range of learning rate hyper-parameters in the grid search, such that the total number of parameter combinations for each condition is approximately the same.

When using a public domain competitive network (\textbf{case 4}), the published learning rate scheduling is used. Therefore we employ the \emph{varied exponential pacing function} without additional learning rate tuning and perform a coarse grid search on the \emph{pacing} hyper-parameters. To ensure a fair comparison, we repeat the experiment with the \emph{vanilla} condition the same number of times as in the total number of experiments done during grid search, choosing the best results. The exact range of values that are used for each parameter is given below in \app~\ref{app:hyper-parameters-ranges}. All prototypical results were confirmed with cross-validation, showing similar qualitative behavior as when using the coarse grid search.

\subsection*{Learning Rate Tuning}
\label{app:learning-rate-tuning}
\begin{figure}[ht!]
    \centering
    \begin{subfigure}[t]{0.49\textwidth}
        \includegraphics[width=\textwidth]{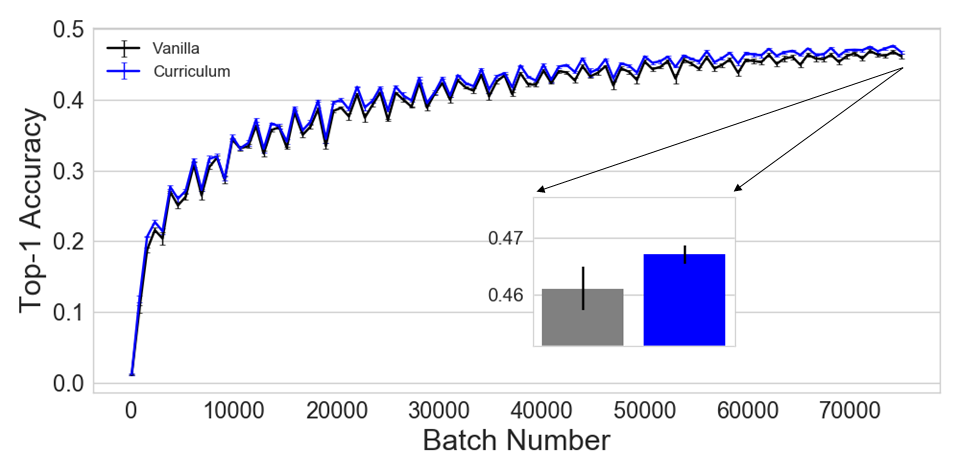}
        %\subcaption{Our results}
        %\label{fig:problem1-ours}
    \end{subfigure}
    \caption{Results under conditions similar to test \textbf{case 3} as shown in \ref{fig:[problem23]}, using cyclic scheduling for the learning rate as proposed by \citet{smith2017cyclical}.}
    \label{fig:appendix-cycle}
\end{figure}

To control for the possibility that the results we report are an artifact of the way the learning rate is being scheduled, which is indeed the method in common use, we test other learning rate scheduling methods, and specifically the method proposed by \citet{smith2017cyclical} which dynamically changes the learning rate, increasing and decreasing it periodically in a cyclic manner. We have implemented and tested this method using \textbf{cases 2} and \textbf{3}. The final results of both the \textsl{vanilla} and \textsl{curriculum} conditions have improved, suggesting that this method is superior to the na\"ive exponential decrease with grid search. Still, the main qualitative advantage of the CL algorithm holds now as well - CL improves the training accuracy during all stages of learning. As before, the improvement is more significant when the training dataset is harder. Results for \textbf{case 3} (CIFAR-100) are shown in Fig.~\ref{fig:appendix-cycle}.

\section{Methodology, additional details}
\label{app: addition-details}
\paragraph{Exponential Pacing}
Throughout this work, we use \emph{pacing functions} that increase the data size each \textsl{step} exponentially. This is done in line with the customary change of learning rate in an exponential manner. 

\paragraph{Architecture Details}
\label{app:architecture}

The moderate-size neural network we used for \textbf{cases 1,2,3,6}, is a convolutional neural network, containing 8 convolutional layers with 32, 32, 64, 64, 128, 128, 256, 256 filters respectively. The first 6 layers have filters of size $3\times3$, and the last 2 layers have filters of size $2\times2$. Every second layer there is a $2\times2$ max-pooling layer and a $0.25$ dropout layer. After the convolutional layers, the units are flattened, and there is a fully-connected layer with 512 units followed by $0.5$ dropout layer. The batch size was 100. The output layer is a fully connected layer with output units matching the number of classes in the dataset, followed by a softmax layer. We trained the network using the SGD optimizer, with cross-entropy loss. All the code will be published upon acceptance.

\paragraph{Grid-search hyper-parameters}
\label{app:hyper-parameters-ranges}
When using grid search, identical ranges of values are used for the \emph{curriculum, anti-curriculum} and \emph{random} test conditions. Since \emph{vanilla} contains fewer parameters to tune -- as it has no \emph{pacing} parameters -- we used a finer and broader search range. The range of parameters was similar between different \emph{scoring functions} and \emph{pacing functions} and was determined by the architecture and dataset. The range of parameters for \textbf{case 1}: (i) initial learning rate: $0.1 \sim 0.01$; (ii) learning rate exponential decrease $2 \sim 1.1$; (iii) learning rate \textsl{step size} $200\sim 800$; (iv) \textsl{step size} $20\sim 400$, for both varied and fixed; (v) \textsl{increase} $1.1 \sim 3$; (vi) \textsl{starting percent} $4\%\sim 15\%$ (note that $4\%$ is in the size of a single mini-batch). For \textbf{cases 2, 3} the ranges is wider since the dataset is larger: (i) initial learning rate: $0.2 \sim 0.05$; (ii) learning rate exponential decrease $2 \sim 1.1$; (iii) learning rate \textsl{step size} $200\sim 800$; (iv) \textsl{step size} $100\sim 2000$, for both varied and fixed; (v) \textsl{increase} $1.1 \sim 3$; (vi) \textsl{starting percent} $0.4\%\sim 15\%$. For \textbf{cases 4, 5}, the learning rate parameters are left as publicly determined, while the initial learning rate has been decreased by $10\%$ from $0.1$ to $0.09$. The \emph{pacing} parameter ranges are: (i) \textsl{step size} $50\sim 2500$, for both varied and fixed; (ii) \textsl{increase} $1.1 \sim 2$; (iii) \textsl{starting percent} $2\%\sim 20\%$. For \textbf{case 6}: (i) initial learning rate: $0.2 \sim 0.01$; (ii) learning rate exponential decrease $3 \sim 1.05$; (iii) learning rate \textsl{step size} $300\sim 5000$; (iv) \textsl{step size} $50\sim 400$; (v) \textsl{increase} $1.9$; (vi) \textsl{starting percent} $2\%\sim 15\%$.

\paragraph{ImageNet Dataset Details}
\label{app: imagenet-cats}
In \textbf{case 6}, we used a subset of the ImageNet dataset ILSVRC 2012. We used 7 classes of cats, which obtained by picking all the hyponyms of the cat synset that appeared in the dataset. The 7 cat classes were: 'Egyptian cat', 'Persian cat', 'cougar, puma, catamount, mountain lion, painter, panther, Felis concolor', 'tiger cat', 'Siamese cat, Siamese', 'tabby, tabby cat', 'lynx, catamount'. All images were resized to size $56\times 56$ for faster performance. All classes contained 1300 train images and 50 test images. The dataset mean was normalized to 0 mean and STD 1 for each channel separately.

\paragraph{Robustness Of Results}
\label{app: robustness}
The learning curves are shown in Fig.~\ref{fig:problem1-summary} were obtained by searching for the parameters that maximize the final accuracy. This procedure only takes into account a few data points, which makes it less robust.
In Fig.~\ref{app: fig:problem1-summary-trapz} we plot the bars of the final accuracy of the learning curves obtained by searching for the parameters that maximize the Area Under the Learning Curve. AUC is positively correlated with high final performance while being more robust. Comparing the different conditions using this maximization criterion gives similar qualitative results - the performance in all the curriculum conditions is still significantly higher than the control conditions. However, now the curriculum based on the Inception-based \emph{scoring function} with \emph{fixed exponential pacing} achieves performance that is significantly higher than the other curriculum methods, in evidence that it is more robust.

\paragraph{Theoretical Section}
\label{app: theoretical}
Proof for proposition 2:
\begin{proof}
Claim 1 follows directly from (\ref{eq:assumption}), while for claim 2:
\begin{equation*}
\begin{split}
\gV(\tilde\vartheta)&-\gV(\vartheta)=\gV(\tilde\vartheta)-\gU(\vartheta) - \hat\Cov[U_\vartheta,p] \\
&\geq \gV(\tilde\vartheta)-\gU(\vartheta) - \hat\Cov[U_{\tilde{\vartheta}},p] = \gU(\tilde\vartheta)-\gU(\vartheta)
\end{split}
\end{equation*}
\end{proof}

\bibliography{curriculum}
\bibliographystyle{icml2019}

\end{document}

%% file: math_commands.tex
\usepackage{amsmath,amsfonts,bm}

\def\eqref#1{equation~\ref{#1}}
\def\ceil#1{\lceil #1 \rceil}
\def\floor#1{\lfloor #1 \rfloor}
\def\1{\bm{1}}

\def\vp{{\bm{p}}}

\def\vx{{\bm{x}}}

\DeclareMathAlphabet{\mathsfit}{\encodingdefault}{\sfdefault}{m}{sl}
\SetMathAlphabet{\mathsfit}{bold}{\encodingdefault}{\sfdefault}{bx}{n}

\def\gH{{\mathcal{H}}}

\def\gU{{\mathcal{U}}}
\def\gV{{\mathcal{U}_p}}

\def\sB{{\mathbb{B}}}

\def\sR{{\mathbb{R}}}

\def\sX{{\mathbb{X}}}

\newcommand{\E}{\mathbb{E}}
\newcommand{\Ls}{\mathcal{L}}

\newcommand{\Var}{\mathrm{Var}}

\newcommand{\Cov}{\mathrm{Cov}}
\DeclareMathOperator*{\argmax}{arg\,max}
\DeclareMathOperator*{\argmin}{arg\,min}